\documentclass[a4paper,11pt]{article}

\usepackage[utf8]{inputenc}
\usepackage[LGR,T1]{fontenc}
\usepackage{gentium}

\advance \topmargin by -\headheight
\advance \topmargin by -\headsep

\usepackage{varwidth}
\usepackage{xstring}
\usepackage{array}
\usepackage{calculator}
\usepackage[hyphens]{url}
\usepackage{relsize} 
\usepackage{graphicx}
\usepackage{amssymb}
\usepackage{comment}
\usepackage{amsmath}

\numberwithin{equation}{section}

\usepackage[style=ieee, 
citestyle=ieee]{biblatex}
\bibliography{references}

\providecommand{\keywords}[1]
{
  \small	
  \textbf{\textit{Keywords---}} #1
}

\title{Class-Wise Principal Component Analysis for hyperspectral image feature extraction}
\author{
  Dimitra Koumoutsou\\
  \texttt{dkoumoutsou@gmail.com}
    \and
  Eleni Charou\\
  \texttt{exarou@iit.demokritos.gr}
    \and \\
  Georgios Siolas\\
  \texttt{gsiolas@islab.ntua.gr}
    \and \\
  Giorgos Stamou\\
  \texttt{gstam@cs.ntua.gr}
}
\date{}

\begin{document}

\maketitle

\begin{abstract}
     This paper introduces the Class-wise Principal Component Analysis, a supervised feature extraction method for hyperspectral data. Hyperspectral Imaging (HSI) has appeared in various fields in recent years, including Remote Sensing. Realizing that information extraction tasks for hyperspectral images are burdened by data-specific issues, we identify and address two major problems. Those are the Curse of Dimensionality which occurs due to the high-volume of the data cube and the class imbalance problem which is common in hyperspectral datasets. Dimensionality reduction is an essential preprocessing step to complement a hyperspectral image classification task. Therefore, we propose a feature extraction algorithm for dimensionality reduction, based on Principal Component Analysis (PCA). Evaluations are carried out on the Indian Pines dataset to demonstrate that significant improvements are achieved when using the reduced data in a classification task. 
\end{abstract}

\keywords{Principal Component Analysis, feature extraction, class imbalance, hyperspectral image, dimensionality reduction}

\normalsize

\section{Introduction}

\emph{Hyperspectral imaging} is essentially the recording of a "scene" in multiple spectra, ranging from the visible to the Near Infra-Red (NIR) \cite{shaw2002signal}. Hyperspectral images are comprised of continuous and contiguous spectral bands at hundreds of narrow wavelength intervals. The resulting image is what we call “hyperspectral scene”, which is the depiction of a geographic location that contains both spectral and spatial information. Especially in the spectral domain, the resolution of the information is very high. Due to this fact, hyperspectral data offer great potential, therefore robust processing strategies need to be employed for maximizing information extraction. 

\begin{figure}[t] \centering
        \includegraphics[width = 0.25\linewidth]{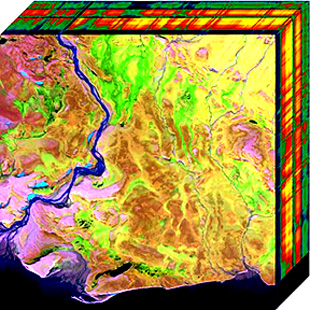} \caption{The hyperspectral cube \cite{wikipedia2021cube}. The first 2 dimensions correspond to the spatial information, while the third dimension is the spectral dimension}\label{figureCUBEen}
        \end{figure}

The data recorded from a Hyperspectral sensor have a 3-dimensional structure, which is called \emph{data cube} \cite{bioucas2013hyperspectral}, as shown in Figure~\ref{figureCUBEen}. The first 2 dimensions correspond to the spatial information, while the third dimension is the spectral dimension. Hyperspectral data have a high resolution in both the spectral and spatial domain. Because they contain spatial information for hundreds of contiguous spectral bands, the hyperspectral cube is a high-volume data cube.

\emph{Class imbalance} in a dataset is a common problem problem which can be described formally as an unequal distribution of samples per class label \cite{japkowicz2000class}. Classes with fewer samples are likely to be under-represented in a classification model, affecting its accuracy. There is also the possibility that classes with few samples are more important for an application depending on the desired task than their statistical insignificance suggests. In this case, the model’s accuracy may seem acceptable, because all the samples play an equal role, even if the majority belongs to one or few classes. The minority of the “weak” class samples hardly affect the accuracy metric, making it misguiding for evaluation. 

The \emph{Curse of Dimensionality}, or Hughes effect, was first mentioned by Richard Bellman in 1961 in his work in Dynamic Programming, as the term used to describe the issue of “an increase in sparseness and dissimilarity of data” which occurs as dimensionality increases \cite{bellman2015adaptive}. Although it would seem that denser information collected by sensors should result in more accurate classification outcomes, this is not always the case with hyperspectral data. The problem is that an increase in dimensionality will actually deteriorate classification, affecting clustering, and the categorization of objects, materials or targets in general. Large amounts of data do not always coincide with a large amount of information; in fact, useful information is not present in every data point of the hyperspectral cube. Information is sparse and because spectral bands are recorded at very narrow wavelength intervals, neighboring bands are most likely correlated. Therefore, using the original recorded data cube does not only slow down the information extraction process; it actually deteriorates it. This is partly because a significant amount of data in the recording is actually noise, but also because training a neural network with redundant data often leads to the model overfitting the training data.

\emph{Dimensionality Reduction (DR)} \cite{van2009dimensionality} is the branch of mathematics used to harvest the vast amount of information stored in hyperspectral data, while dealing with the Curse of Dimensionality issue. Dimensionality Estimation and Reduction is the process of transforming a cube with $(X, Y)$ spatial dimensions, and $L$ spectral dimensions, represented as a $3-$D vector as $(H, X, L)$, to a lower-dimensional cube $(H, Y, N)$, where $N<L$. 
Since many channels in a hyperspectral image are correlated and/or noisy, redundant data could be eliminated from datasets. The challenge of Dimensionality Reduction is to dispose of unnecessary or low quality information, while retaining the total of the significant features about the targets of interest. In Remote Sensing HSI applications Dimensionality Reduction is used as a preprocessing step, mainly in two main forms, band selection \cite{sun2019hyperspectral} or feature extraction \cite{zhao2016spectral}. So far, literature findings indicate that the latter is the optimal approach and is an indispensable preprocessing step when analyzing hyperspectral data. A powerful feature extraction algorithm should retain only useful information and discard all redundancies in the dataset, so implementing it as a preprocessing step in a classification model is expected to reduce computational time and improve overall performance. 

\section{Principal Component Analysis (PCA)}

Principal Component Analysis (PCA) \cite{pearson1901liii} is also called Karhunen–Loève transform. It is a powerful technique of multivariate analysis which seeks for the projection that best represents data in the least square sense. PCA is commonly used for dimensionality reduction in an automatic fashion. This method is based on rotating the original data into a set of axes that maximizes the variability in the first few axes. This is done by finding a new coordinate system in the hyper-dimensional vector space where the data exhibits no correlation \cite{shlens2014tutorial}. Let, 
        \begin{gather}
            X_i = [x_1, x_2, ...x_N]_i^T \\
            m = \frac{1}{M}\displaystyle\sum\limits_{i=1}^M  [x_1, x_2, ...x_N]_i^T \\
            C_x = \frac{1}{M} \displaystyle\sum\limits_{i=1}^M  (x_i - M)(x_i - M)^T 
        \end{gather}
Where, \\
$X_i$ is a hyperspectral image vector of all $N$ dimensions of a given pixel, located at $i$, \\ $x_i$ is a pixel at the i-th spectral band, \\
$M$ is the total number of pixels in each band,\\
$m$ is the mean value of all the pixels in the image, \\
$N$ is the number of spectral dimensions in the original data, \\
$C_x$ is the covariance matrix, which is diagonal or uncorrelated in the new coordinate system. \\

A linear transformation, $U$, that transforms the original hyperspectral data, $X_i$, into the new coordinate system $Y$ is then  calculated. The original covariance matrix $C_x$ becomes a diagonalized covariance matrix. The solution becomes a generalized eigenvalue problem of the form:
    \begin{equation}
        C_x = UDU^T 
    \end{equation}
where
\[D = diag(\lambda_1, \lambda_2, .., \lambda_N) \]
\[U = (u_1, u_2, .., u_N) \]
Here, \\
$D$ is the diagonal matrix of $\lambda_I$\\
$\lambda_I$ are the eigen-values of $C_x$\\
$C_x$ is the covariance matrix\\
$U$ is the orthogonal matrix, which is the matrix of $u_i$\\
$u_i$ are the eigenvectors.\\

Finally, the PCA transform is computed by 
    \begin{equation}
        Y_i = U^TX_i 
    \end{equation}

where \[i = 1, 2, \dots, M\] 
Each of the vectors $Y_i$ contains the reduced information of the original hyperspectral image. The number of the components used in this final step depends on the intended information extraction task and will be thoroughly investigated in our experiments section. 
\section{Proposed Method}

We introduce \emph{{Class-Wise Principal Component Analysis (CW-PCA)}}, which is a supervised alternative to the classic PCA algorithm described in the previous Section. 

\begin{figure}[t] \centering
        \includegraphics[width=\linewidth]{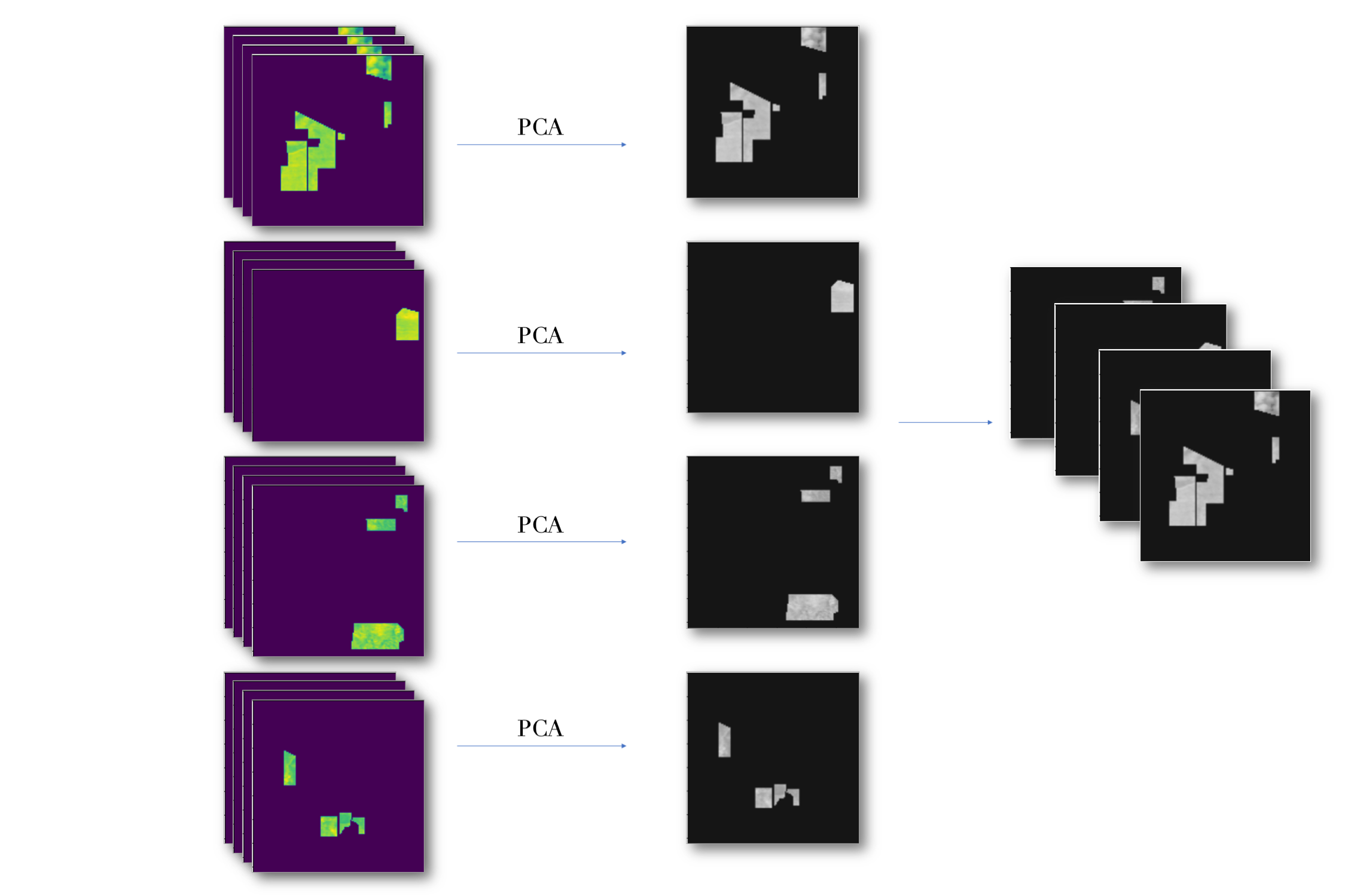} \caption{Outline for the proposed feature extraction technique, Class-Wise Principcal Component Analysis (CW-PCA).}\label{figureCWPCA}
        \end{figure}

The motivation behind this new approach is that the features that contain most of the variance are not necessarily those that contain the information we wish to preserve. Besides, we should further consider the issue of class imbalance. The classes that are already ill-represented in the original space are prone to further elimination in the transformed subspace. The challenge that should be addressed is the following; if the spectral information that is most informative about a class is discarded during the dimensionality reduction process, there is a risk of the class being misclassified. Although this could occur both for “strong” and “weak” classes, it is only the weak ones that would suffer the greatest hit. Since those classes are not predominant, they heavily rely on the abundance of spectral features and the ability of a fit classifier to recognize their unique spectral signatures. If that spectral signature is altered in a way that discards the uniqueness of the object/material, nothing else can be used to their advantage. On that account, extracting principal components from the entire dataset poses a risk. We wish to find a method to drastically reduce dimensionality with minimal loss of information and maximal discarding of redundancy. To ensure that the low-dimensional subspace bears all of the important features of the original dataset, we introduce the concept of CW-PCA, which is outlined in Figure~\ref{figureCWPCA}. 

Let $(X, Y, L)$ be the dimensions of the original image where $(X, Y)$ denote the spatial information and $C$ is the number of bands. Assuming there is a known number of classes in the dataset ($N-$ classes), we may reduce dimensionality from $L$ to a multiple of $N$. To implement this method, we first split the original scene into $N$ subscenes. Each of them has the same dimensions $(X, Y, L)$ as the original scene, but only has positive values in the coordinates of the $N-$th class samples. Essentially, we create $N$ non-overlapping subscenes each of whom only carries samples from one class. Then, PCA is applied separately on each of these subscenes and one (or more) Principal Components are extracted from each subscene. At the final step, the Principal Components (PCs) are concatenated along the spectral dimension. For example, if only the first PC is kept from each class-scene, at this point we should have $N$ Principal Components, each with dimensions $(X, Y, 1)$, and the concatenated vector-subspace would be $(X,Y,N)$ dimensional. It is evident from the described process that the number of PCs in the resulting subspace should be a multiple of the number of classes. More specifically, if the $m-$th first PCs are selected from each class, the resulting spectral dimension is $m\cdot N$.
\section{Experiments}

Four feature extraction methods are tested, three of which are already used in hyperspectral imaging for dimensionality reduction. Then, we propose a new method, Class-wise PCA, which we will examine in more depth to determine whether it can actually contribute to resolving the two problems we discussed: dimensionality reduction and class imbalance. Although class imbalance can be solved at data-level (for example with data augmenation techniques \cite{shorten2019survey}), we want to investigate if using a supervised method to incorporate class information at the feature extraction stage can have an effect on classification accuracy, especially in class-wise accuracy. 

To rank the three methods against each other and evaluate our proposed method accordingly, we use all four different methods to reduce dimensionality and perform classification on the data. So far as unsupervised methods are concernes, experimental results prove that 15 PCA components are enough to convey the information needed to achieve maximum accuracy. On the other hand, when transforming data with a supervised approach, class information is considered in the transformed lower-dimensional  space. In the case of LDA this also means that, at most, the number of bands in the transformed space can be equal to the number of classes. As previously stated, CW-PCA transforms data to a number of bands that is an integer multiple of the number of classes. Holding on to the same reasoning, we will keep one component from each class in the CW-PCA algorithm, which will give us equal number of bands as the LDA algorithm. Since there are 16 classes in our dataset, and we found the optimal number of components in the unsupervised algorithms to be 15, we can have a fair comparison across the four methods as the reduction in dimensionality is roughly the same. 

\subsection{Dataset}
The Indian Pines dataset is a publically available hyperspectral scene \cite{IPdataset}. It  consists of $145\times145$ pixels in the spatial dimension and 224 spectral bands, although the corrected version we use has $220$, in the range of $400$ to $2500nm$. Data was captured with the Airborne Visible/Infrared Imaging Spectrometer (AVIRIS) sensor \cite{green1998imaging} in North-Western Indiana, USA, with a Ground Sample Distance (GSD) of $20m$ and are labeled with 16 classes of vegetation not all of which are mutually exclusive; more specifically there are two-thirds agriculture, and one-third forest or other natural vegetation. The 16 labeled  classes are
\begin{quotation}
\noindent'Alfalfa',  'Corn-notill',  'Corn-mintill',  'Corn', 'Grass-pasture', 'Grass-trees', 'Grass-pasture-mowed', 'Hay-windrowed', 'Oats', 'Soybean-notill', 'Soybean-mintill', 'Soybean-clean', 'Wheat', 'Woods', 'Buildings-Grass-Trees-Drives', 'Stone-Steel-Towers'
\end{quotation}

    \begin{figure}[t]
    \centering
          \centering
          \includegraphics[width=0.49\linewidth]{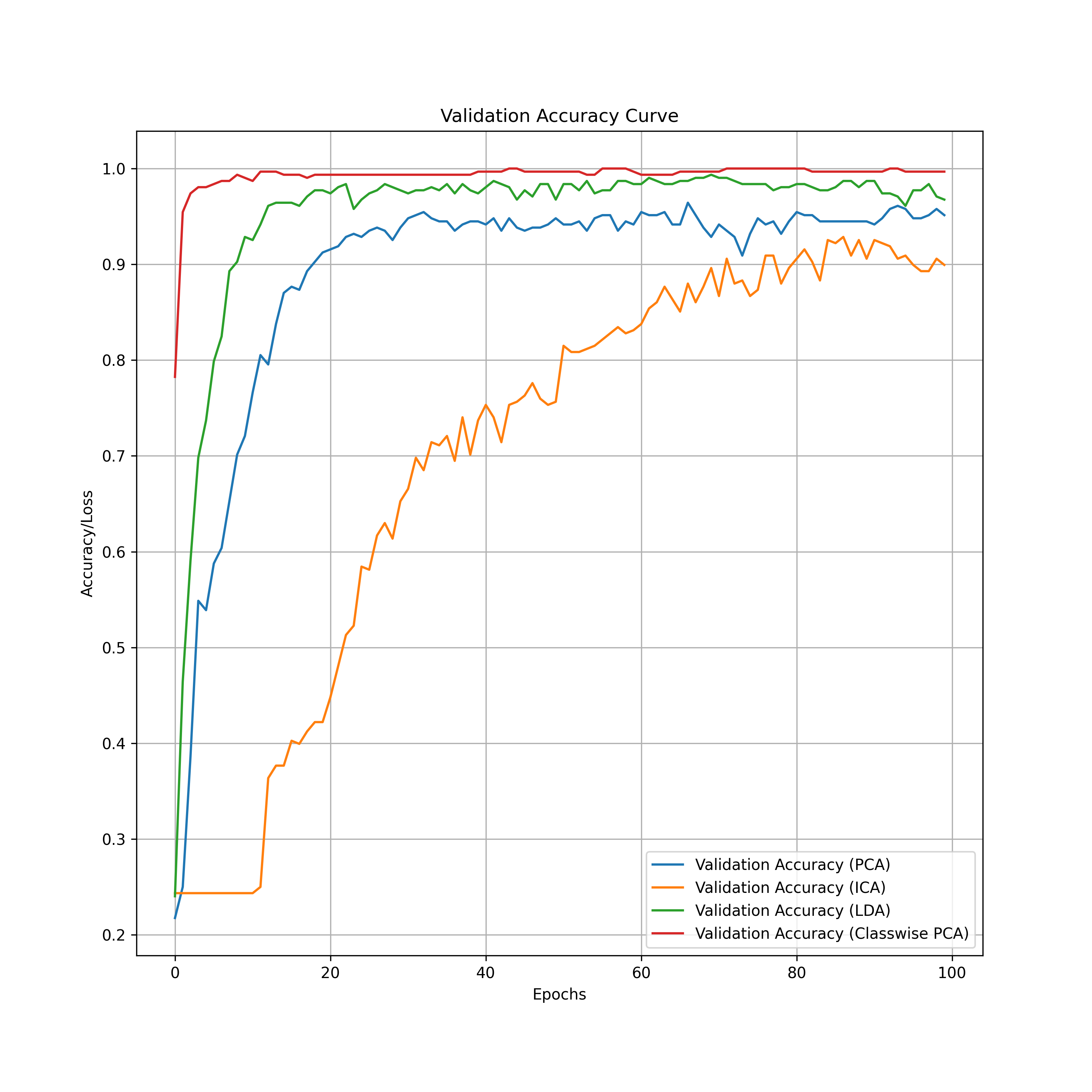}
          \centering
          \includegraphics[width=0.49\linewidth]{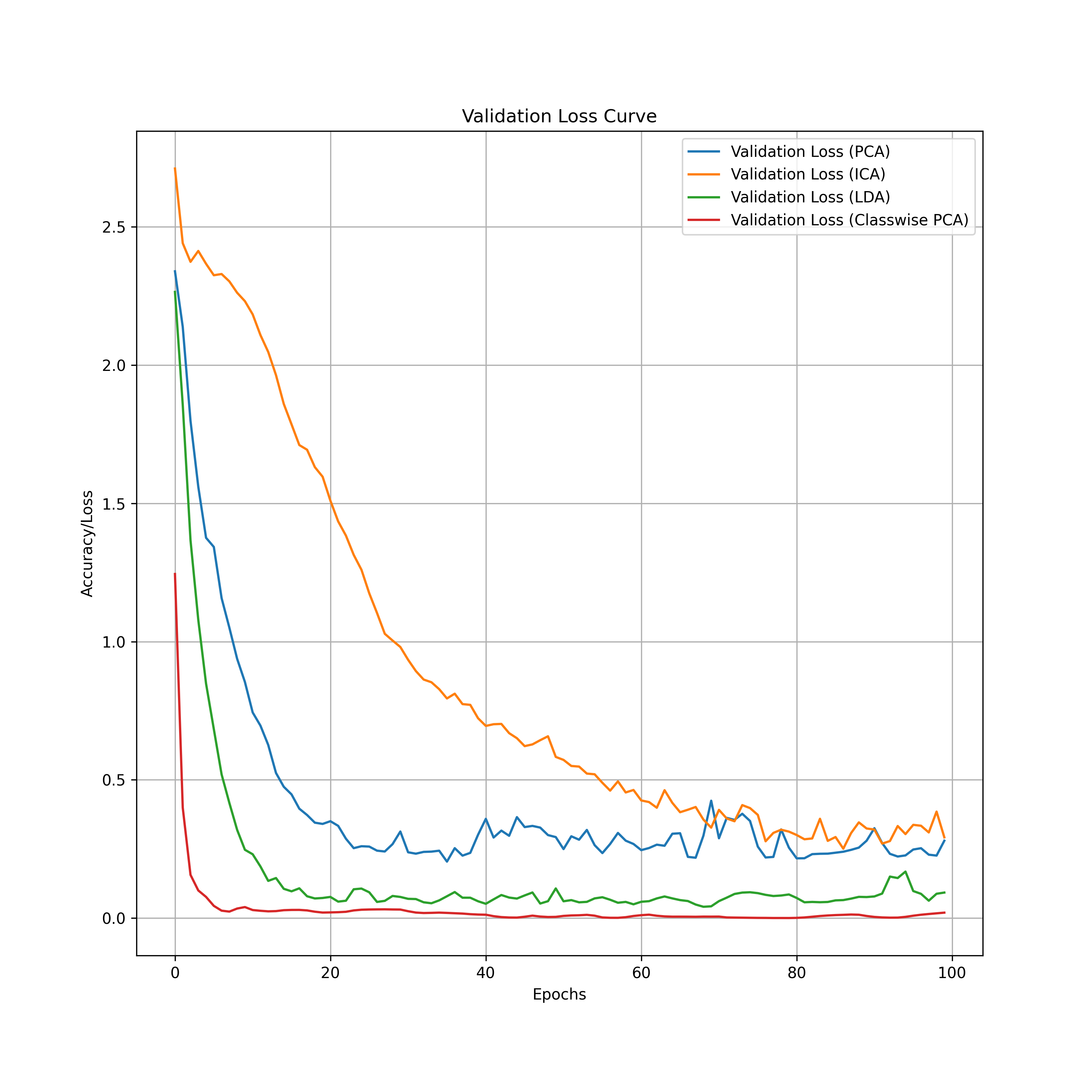}
        \caption{(a) Accuracy and (b) Loss curves for the training and validation set, when performing classification on data reduced by PCA, ICA, LDA and CW-PCA. The CW-PCA method converges faster and shows no risk of overfitting.}
    \label{figureDRaccuracyALL}\end{figure}
    
\subsection{Dimensionality Reduction}
The first aspect is the dimensionality reduction. Essentially, we want to determine if reducing the spectral dimensions by up to 90\%, from the original 220 bands to only 16 (for the Indian Pines dataset), will accurately convey all useful information and discard redundancies. This will be revealed in the classification outcomes in two ways. If the dimensions are not effectively reduced, the redundant information will introduce noise to the model and skew the results. In contrary, if we shrink the input too much, we may have extreme loss of information which will simply not suffice to train a reliable model. We can see how each methods performs in terms of classification accuracy in Figure~\ref{figureDRaccuracyALL}(a), while the loss curve for the train and validation sets is given in Figure~\ref{figureDRaccuracyALL}(b). ICA has the worst performance out of the four, both in terms of convergence speed (over 70 epochs) and accuracy. The PCA and LDA are similar, with the latter outperforming PCA slightly. Hence it appears that including class information in the preprocessing steps is of essense. Finally, the CW-PCA has an outstanding performance, with very fast convergence in 10 epochs, which is 50\% faster than PCA and LDA. Also, from the loss curve we see that even after 90 training epochs, there is not a slightest hint for overfitting. 
A quantitative evaluation of the classification scores is given in Table~\ref{tableDRaccuracy}, which corroborates the quantitative information.  
        	\begin{table}[ht]
    		\centering
    		\small
    		\renewcommand{\arraystretch}{1.3}
    		\begin{tabular}{| c || c | c |}
    		 \hline               
    		  DR Method & OA & AA \\
    		   \hline
    			    ICA & 89.43 & 81.48\\
    			    PCA & 91.53 & 84.79\\
    			    LDA & 96.43 & 90.12\\
    			    \textbf{CW-PCA} & \textbf{99.95} & \textbf{99.82}\\
    			\hline
    		\end{tabular}
    		\caption{Classification accuracy scores (OA and AA) for different methods of feature extraction (PCA, ICA, LDA, CW-PCA)}.
    		\label{tableDRaccuracy}
    	\end{table}

    \begin{figure}[ht] \centering
    \includegraphics[width=0.7\linewidth]{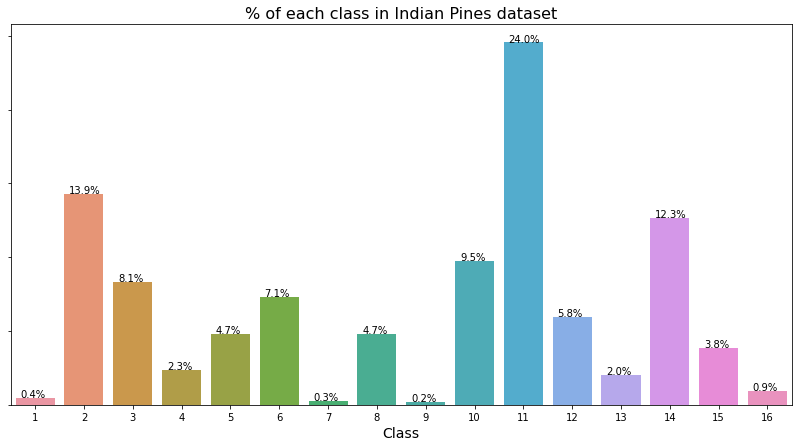} \caption{Percentages of training samples per class for the Indian Pines Dataset. We observe the magnitude of the class imbalance problem.}\label{figureIMBALANCEen}
    \end{figure}

\subsection{Class imbalance}

In Figure~\ref{figureIMBALANCEen} we observe the significant class imbalance problem for the Indian Pines Dataset, which is a benchmark dataset for the majority of hyperspectral image classification models in the literature. Next we evaluate the CW-PCA method against the other three algorithms with respect to their contribution in solving the class imbalance problem. Although the improvement is already evident from the Average Accuracy metric in Table~\ref{tableDRaccuracy}, which expressed the class-wise accuracy of our model, we will proceed with some visual results that highlight the superiority of CW-PCA against other methods. This indicates that the high accuracy scores are not a result of few classes dominating the results and misguiding the evaluation scheme. 

\begin{figure}[ht]
        \centering
          \centering
          \includegraphics[width=0.5\linewidth]{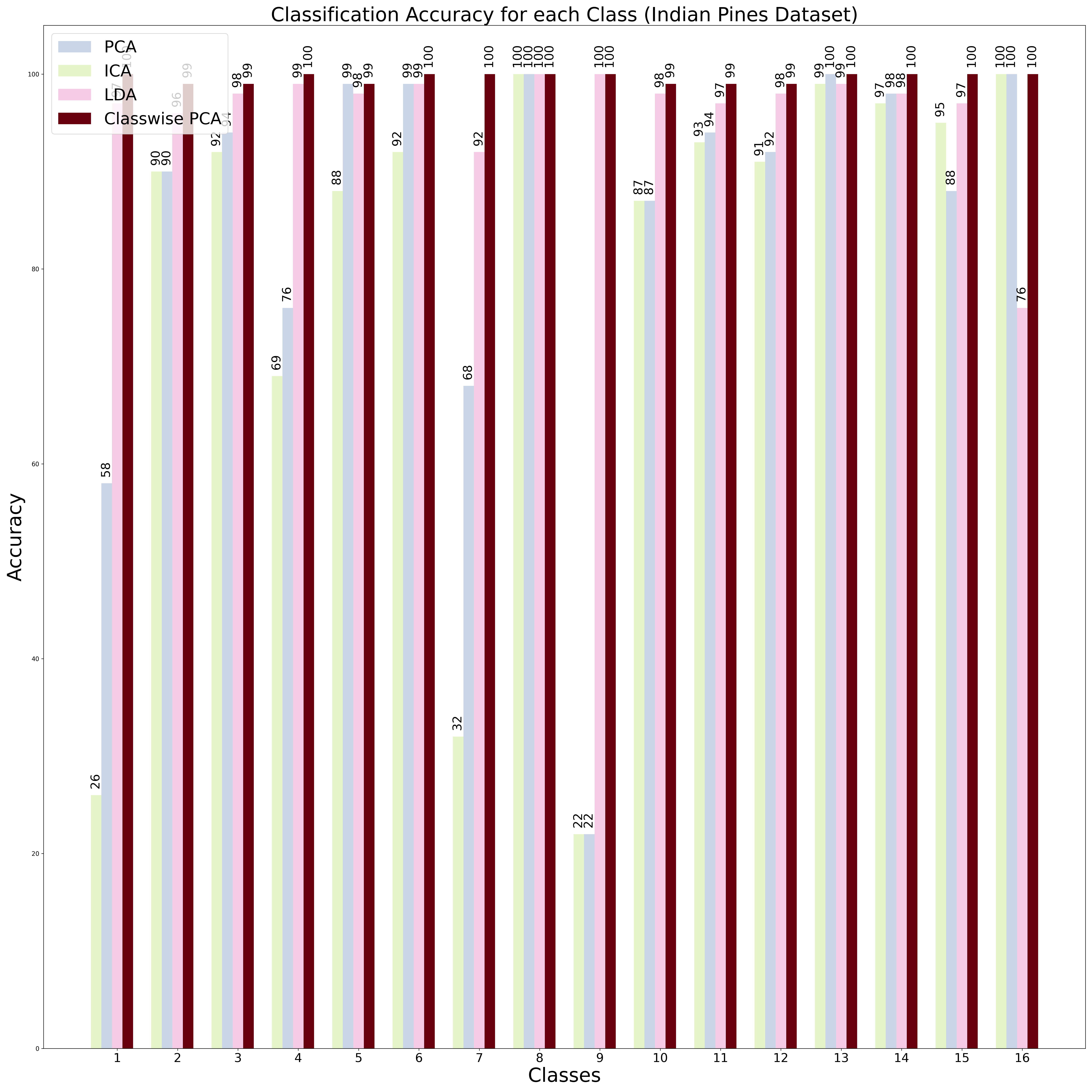} 
           \includegraphics[width=0.5\linewidth]{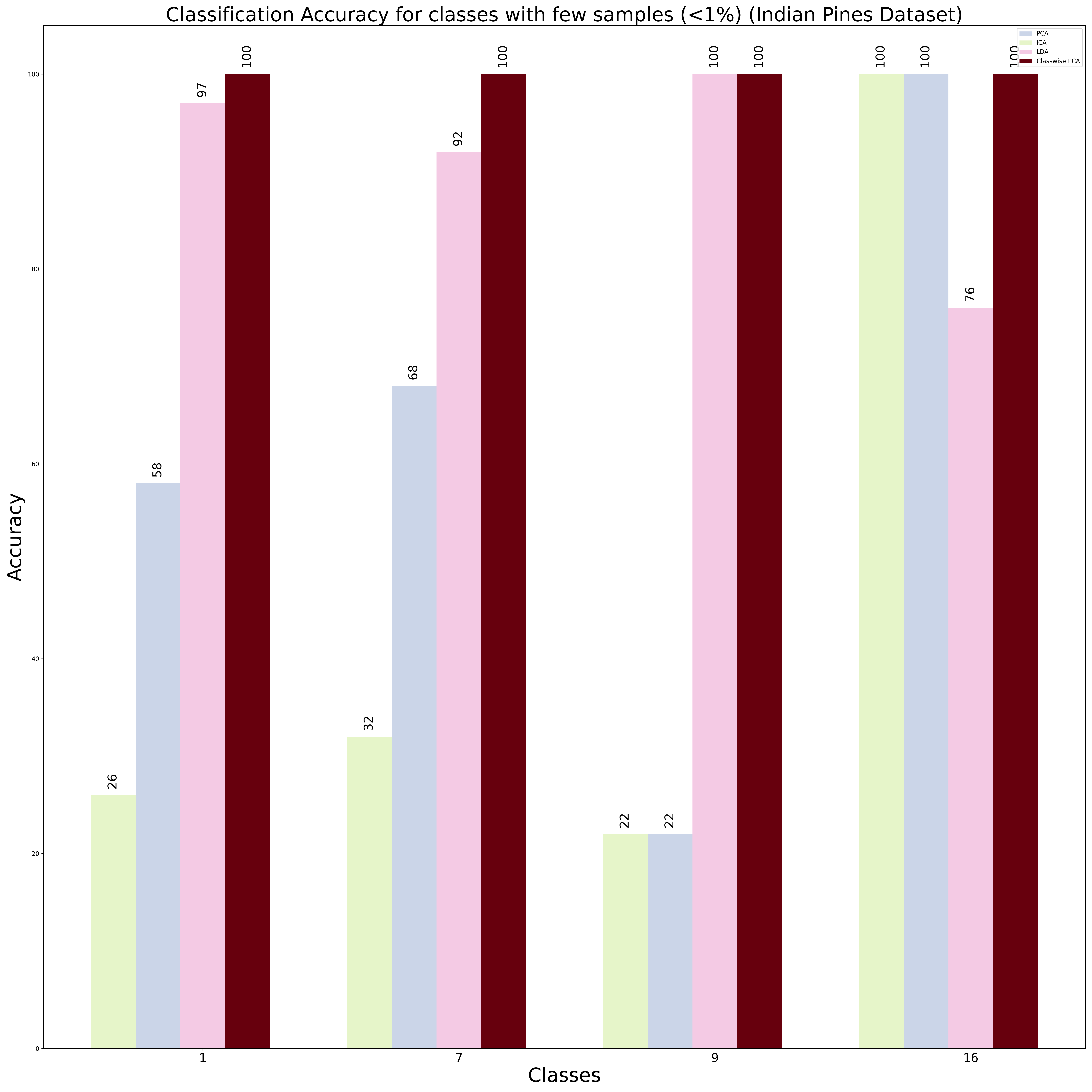}
        \caption{(top) Class-wise accuracy using CW-PCA for dimensionality reduction versus PCA, ICA, LDA. Clearly CW-PCA outperforms all other methods, (bottom) Class-wise accuracy for weak classes of the Indian Pines Dataset, which have less than 1\% of total training samples (classes 1, 7, 9, 16). Comparison of using CW-PCA for dimensionality reduction versus PCA, ICA, LDA.}
    \label{figureCWaccALL}\end{figure}

To further investigate the issue of class imbalance using the class-wise accuracy, we plot the accuracy per class. As can be seen in the class imbalance plot in Figure~\ref{figureIMBALANCEen}, the weak classes for the Indian Pines dataset are 1, 7, 9 and 16. In Figure~\ref{figureCWaccALL} we see that CW-PCA has almost 100\% accuracy even in the case of weak classes, while other methods yield significantly poorer results. LDA seems to perform better at weak classes than the unsupervised counterparts, however there are still inconsistencies, for example class 16 has a 26\% offset than all other methods. To highlight and better observe how weak classes were classified, we “zoom in” on those four classed for a close-up. The bottom graph in Figure~\ref{figureIMBALANCEen} underlines the improvement in weak classes and proves that CW-PCA outperforms all other methods. Amongst the three other methods (PCA, ICA, LDA), there is no clear ranking. Even if LDA appears better in most cases the results are not always consistent. Consequently, even amongst supervised methods, CW-PCA is the most reliable choice.

\newpage

\newpage
\printbibliography

\end{document}